\documentclass{article}

\usepackage[nonatbib, final]{neurips_2024}

\usepackage[utf8]{inputenc} 
\usepackage[T1]{fontenc}    
\usepackage{hyperref}       
\usepackage{url}            
\usepackage{booktabs}       
\usepackage{amsfonts}       
\usepackage{nicefrac}       
\usepackage{microtype}      
\usepackage{xcolor}         
\usepackage{graphicx}
\usepackage{subfigure}
\usepackage[numbers]{natbib}
\usepackage{amsmath}
\usepackage{multirow}
\usepackage{wrapfig}
\usepackage{float}
\usepackage{amssymb}
\usepackage{colortbl}

\definecolor{beaublue}{rgb}{0.74, 0.83, 0.9}
\title{HYDRA-FL: Hybrid Knowledge Distillation for Robust and Accurate Federated Learning}

\author{
  Momin Ahmad Khan\\
  University of Massachusetts, Amherst\\
  \texttt{makhan@umass.edu} \\
  \And
  Yasra Chandio \\
  University of Massachusetts, Amherst \\
  \texttt{ychandio@umass.edu} \\
  \AND
  Fatima Muhammad Anwar \\
  University of Massachsuetts, Amherst \\
  \texttt{fanwar@umass.edu} \\
}

\begin{document}

\maketitle

\begin{abstract}
    Data heterogeneity among Federated Learning (FL) users poses a significant challenge, resulting in reduced global model performance. The community has designed various techniques to tackle this issue, among which Knowledge Distillation (KD)-based techniques are common.
    While these techniques effectively improve performance under high heterogeneity, they inadvertently cause higher accuracy degradation under model poisoning attacks (known as \emph{attack amplification}). This paper presents a case study to reveal this critical vulnerability in KD-based FL systems. We show why KD causes this issue through empirical evidence and use it as motivation to design a hybrid distillation technique. We introduce a novel algorithm, \emph{Hybrid Knowledge Distillation for Robust and Accurate FL (HYDRA-FL)},
    \footnote{We will release the open source code with the final version of this paper.},
    which reduces the impact of attacks in attack scenarios by offloading some of the KD loss to a shallow layer via an auxiliary classifier. We model HYDRA-FL as a generic framework and adapt it to two KD-based FL algorithms, FedNTD and MOON. Using these two as case studies, we demonstrate that our technique outperforms baselines in attack settings while maintaining comparable performance in benign settings. 
\end{abstract}
\section{Introduction}\label{sec:introduction}
Federated Learning (FL)~\cite{mcmahan2017communication} is an emerging machine learning paradigm enabling multiple users' collaborative model training without data sharing. Each user, termed a \emph{client}, only shares their local model with a \emph{server}, which aggregates all local models into a single global model and redistributes it to the clients. 
Due to its decentralized, privacy-preserving, and highly-scalable nature, FL has been adopted by Google's Gboard~\cite{gboard} for next-word prediction, Apple's Siri~\cite{technologyreviewApplePersonalizes} for automatic speech recognition, and WeBank~\cite{webankcredit} for credit risk prediction.

Despite its benefits, FL faces challenges with data heterogeneity~\cite{li2019convergence, zhao2018federated, hsu2019measuring, li2022federated}. FL performs well when client data is independent and identically distributed (IID) and achieves similar convergence as a single model trained on all the clients' data but struggles when clients have diverse data (non-IID). In this case, the client's local data is not a good representation of the overall data distribution (unlike an ideal IID case), causing local models to \emph{drift} away from each other. This drift results in a global model with significant accuracy degradation compared to the IID scenario. Numerous solutions~\cite{li2021model, lee2022preservation, li2019fedmd, zhu2021data, li2019rsa, xie2022robust, karimireddy2020scaffold, li2020federated} address data heterogeneity, including Knowledge Distillation (KD)~\cite{hinton2015distilling} to reduce the drift between local models. 

Besides data heterogeneity, FL also faces the issue of Byzantine robustness~\cite{kairouz2019advances}, where \emph{untrusted clients} can inject \emph{poisoned} models into the aggregator by altering client data (data poisoning~\cite{munoz2017towards}) or client models (model poisoning~\cite{fang2020local, baruch2019a, mhamdi2018the, xie2019fall, bhagoji2019analyzing, bagdasaryan2018how, shejwalkar2021manipulating}). Research by \cite{shejwalkar2022back} shows that model poisoning attacks are more potent as they directly manipulate local models. To counteract poisoning in FL, various defenses have been developed~\cite{blanchard2017machine, yin2018byzantine, zhang2022fldetector, cao2022fedrecover, li2021ditto, chang2019cronus, cao2021provably}.

In this work, we identify a critical vulnerability in KD-based FL techniques under model poisoning attacks. These techniques unknowingly align benign client models with a poisoned server model (Figure~\ref{fig:problem}). We study two such classes of KD-based solutions: FedNTD~\cite{lee2022preservation}, which reduces the loss between not-true logits of the server and client models, and MOON~\cite{li2021model}, which reduces the contrastive loss between the representation vector of the server and client models. While these techniques improve global model accuracy in benign settings compared to FedAvg~\cite{mcmahan2017communication} (standard FL aggregator), they \emph{reduce performance below FedAvg under attack}, a phenomenon we term \emph{attack amplification}, especially noticeable at higher heterogeneity levels.

Motivated by our findings, we propose a Hybrid Knowledge Distillation for Robust and Accurate FL (HYDRA-FL) framework for KD-based techniques that restricts attack amplification under poisoning attacks while retaining performance in the benign setting. Unlike traditional KD methods that apply KD-loss only at the final layer, HYDRA-FL introduces KD-loss at a \emph{shallow layer} via an auxiliary classifier and reduces the KD-loss impact at the final layer. This approach draws inspiration from Self-Distillation (SD)~\cite{zhang2019your} and Skeptical Students (SS)~\cite{kundu2021analyzing}, but with a distinct focus on enhancing robustness against heterogeneity and model poisoning attacks in FL.

SD improves model accuracy by self-distillation, while SS distills from "nasty teachers"~\cite{ma2021undistillable} to shallow layers. In contrast, our approach uses auxiliary classifiers to enhance FL client robustness against heterogeneity and model poisoning attacks. We design a generic loss function adaptable to specific KD-based algorithms. Extensive experiments show that HYDRA-FL significantly boosts accuracy over FedNTD and MOON in attack settings while maintaining performance in benign settings. 

\textbf{Contributions.} This work addresses the critical issue of attack amplification in KD-based FL techniques to counter data heterogeneity. In doing so we make the following contributions:

\begin{figure}[t]
    \centering
    \includegraphics[width=.9\textwidth]{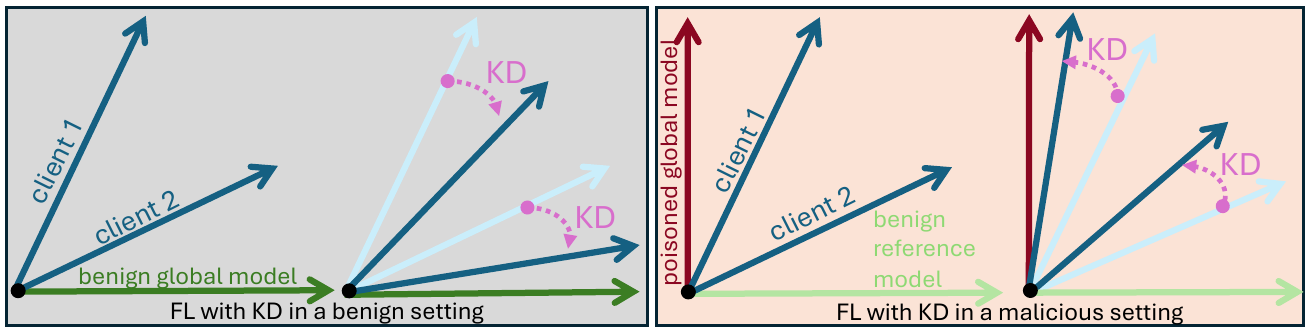}
    \caption{Overview of attack amplification through knowledge distillation. \textbf{a)} In the benign setting, KD reduces drift and brings benign local models closer to the benign global model. \textbf{b)} In the malicious setting, KD \emph{unknowingly} reduces drift between benign local models and the poisoned global model.}
    \label{fig:problem}
\end{figure}
\begin{itemize}
    \item \textbf{Proving KD amplifies model poisoning:} our motivational case study (\S\ref{sec:motivation}) on two KD-based techniques, FedNTD and MOON, shows that KD improves accuracy in benign settings but helps the malicious clients propagate poisoning through the KD-loss in adversarial settings. We empirically and theoretically show that this attack amplification issue is inherent to any technique aligning client outputs/representations with the server.
    \item \textbf{Designing HYDRA-FL:} Using our observations as a guideline, we design HYDRA-FL (\S\ref{sec:solution}) to prevent attack amplification while retaining performance in the benign setting. HYDRA-FL is formulated as a general loss function adaptable to any FL algorithm to use as its local model training objective.
    \item \textbf{Implementation and Evaluation:} we adapt HYDRA-FL to FedNTD and MOON and modify their local training objectives (\S\ref{sec:experiments}). Our qualitative and quantitative analysis (\S\ref{sec:analysis}) shows HYDRA-FL achieves higher accuracy in attack settings and maintains accuracy in benign settings.
\end{itemize}
\section{Background and Related Work}\label{sec:background}

\subsection{Federated Learning (FL)}\label{background:FL}
In FL~\cite{kairouz2019advances, mcmahan2017communication}, a service provider, called \emph{server}, trains a \emph{global model}, $\theta^g$, on the private data from multiple collaborating clients, all without directly collecting their data.
The server selects $n$ out of total $N$ clients in every FL round and shares the most recent global model ($\theta^t_g$) with them, where $t$ is the round number. 
Then, a client $k$ uses their local data $D_k$ to compute an update $\nabla^t_k$ and shares it with the server. The server aggregates these updates using some \emph{aggregation rule}, like FedAvg~\cite{mcmahan2017communication} algorithm.
 
In \emph{FedAvg}, a client $k$ \emph{fine-tunes} $\theta^t_g$ on their local data using stochastic gradient descent (SGD) for a fixed number of local epochs $E$, resulting in an updated local model $\theta^t_k$. The client then computes their update as the difference $\nabla^t_k= \theta^t_k-\theta^t_g$ and shares $\nabla^t_k$ with the server.
Next, the server computes an aggregate of client updates, $f_\mathsf{agg}$ using mean, i.e.,
\begin{equation}
    \nabla^t_\mathsf{agg}= f_\mathsf{mean}(\nabla^t_{\{k\in[n]\}}). 
\end{equation}
 
The server then updates the global model of the $(t+1)^{th}$ round using SGD  and server learning $\eta$ as:
\begin{equation}
    \theta^{t+1}_g\leftarrow \theta^{t}_g+\eta\nabla^t_\mathsf{agg}
\end{equation}

\subsubsection{Data Heterogeneity in FL}
Data heterogeneity is a well-explored problem~\cite{li2019convergence, zhao2018federated, hsu2019measuring, li2022federated} in FL. Each client in FL generates its data, leading to local data distributions that vary across clients and do not accurately represent the global data distribution. By extension, a global model learned by aggregating local models using FedAvg may not be the best representation of all the client's local data. Studies have shown that this data heterogeneity degrades performance and have proposed various methods to address this issue~\cite{li2021model, lee2022preservation, li2019fedmd, zhu2021data, li2019rsa, xie2022robust, karimireddy2020scaffold, li2020federated}. This degradation is more prominent in the presence of poisoning attacks. Research on poisoning attacks in FL has demonstrated that such attacks become more successful under high heterogeneity~\cite{fang2020local, shejwalkar2021manipulating}. This increased risk is because the malicious clients can more easily hide between drifted benign client models, making it difficult for the server to differentiate between heterogeneous benign clients and malicious ones. \cite{khan2023pitfalls} highlights that overlooking this heterogeneity is a critical oversight in FL defense evaluations.

\subsubsection{Poisoning in FL}
FL is vulnerable to poisoning attacks~\cite{blanchard2017machine, baruch2019a, bhagoji2019analyzing, bagdasaryan2018how, mhamdi2018the, fang2020local, mahloujifar2019universal, xie2019fall, munoz2017towards, shejwalkar2021manipulating}, where malicious clients aim to compromise the training process by degrading the global model's performance. These attacks come in various forms: 
In \emph{data poisoning}~\cite{bagdasaryan2018how}, malicious clients poison their local data to introduce a backdoor in the local model. This backdoor then propagates to the global model upon aggregation.
In \emph{model poisoning}~\cite{fang2020local, baruch2019a, mhamdi2018the, xie2019fall, bhagoji2019analyzing, bagdasaryan2018how, shejwalkar2021manipulating}, malicious clients perturb their local models so that, when aggregated, the global model is poisoned.
Poisoning attacks can be further classified based on their targets:
If the performance degradation is on specific inputs, the attack is termed as \emph{targeted poisoning}~\cite{bhagoji2019analyzing, bagdasaryan2018how}, and if it is on all inputs, then it is termed as \emph{untargeted poisoning}~\cite{fang2020local, baruch2019a, mhamdi2018the, mahloujifar2019universal, xie2019fall}. We explain the attacks used in this paper in \S\ref{appdx:adversarial_settings:attacks}.

\subsection{Knowledge Distillation (KD)}\label{background:KD}
Knowledge Distillation (KD)~\cite{hinton2015distilling} transfers knowledge from a large, complex model (\emph{teacher}) to a smaller, more computationally efficient model (\emph{student}). This process involves distilling the teacher's rich and intricate information into the student by aligning their predictions. Formally, if the teacher and student models produce the output probabilities $y^i_t$ and $y^i_s$ respectively for the $i^{th}$ input $(x^i, y^i)$, KD aims to match these probabilities by applying the Kullback-Leibler (KL) divergence between them. The KL-divergence between their softened probabilities is given by:
$KL(softmax(y^i_t/\tau) || softmax(y^i_s/\tau)$, where $\tau$ is the temperature parameter that softens the probabilities.
The overall KD loss function combines this KL-divergence with the usual loss function such as cross-entropy (CE) loss with $\beta$ (balances the importance of the KL-divergence and CE loss) as:
\begin{equation}\label{eqn:KD}
    \mathcal{L} = (1 - \beta)\cdot \mathcal{L}_{CE}(y^i_s, y^i) + \beta\cdot \mathcal{L}_{KL}(softmax(y^i_s/\tau) || softmax(y^i_t/\tau))
\end{equation}

\noindent\textbf{KD in FL} is becoming essential as it addresses critical challenges such as non-IID data distributions, enhances model performance, accelerates convergence, reduces communication overhead, and improves robustness by making the global model learn from an ensemble of local models~\cite{chen2020fedbe, li2019fedmd, lin2020ensemble, zhou2020distilled}. In FL, data is often non-IID across clients, leading to significant discrepancies in local models. KD mitigates these discrepancies by aligning the local models with the global model, ensuring that the global model captures a more generalized representation of the data.
The general approach is to reduce the local model drift by improving the aggregation through distillation using unlabeled auxiliary data. However, the auxiliary data may not always be available, and methods have also been developed to enable KD without such data \cite{zhang2022fine, zhu2021data}.
\section{Attack Amplification through Knowledge Distillation}\label{sec:motivation}
\noindent \textbf{Hypothesis.}
KD-based techniques in FL improve accuracy in non-adversarial settings but result in more significant accuracy degradation under model poisoning attacks compared to the baseline techniques such as FedAvg.

\noindent\textbf{Motivational case study.}
In this case study, we compare FedAvg against two distinct KD-based solutions addressing the local model drift from non-IID. MOON~\cite{li2021model} uses model-contrastive learning to align local and global model \emph{representations}, while FedNTD~\cite{lee2022preservation} uses KL-divergence to align \emph{not-true logits} of client models with those of the server. FedNTD penalizes prediction divergence measured through distillation loss, improving knowledge transfer and stability, while MOON penalizes \emph{representation divergence} measured through contrastive loss, enhancing robustness and generalization. This comparison will help us understand the trade-offs of using KD in FL, especially under adversarial conditions. Throughout this paper, benign conditions mean that no attacks are present, while adversarial conditions mean that model poisoning attacks are present. We implement the same settings and hyperparameters for FedAvg as for MOON and FedNTD to ensure a fair comparison, so FedAvg results may vary between these techniques. This is not an inconsistency.
\emph{We do not directly compare FedNTD to MOON unless stated otherwise}, as the original FedNTD work already did so.
Our goal is to test how adversarial settings affect these two fundamentally different techniques similarly, demonstrating that \emph{attack amplification is inherent to KD and not specific to a particular technique.}

\begin{wrapfigure}{r}{0.6\textwidth}
\vspace{-2em}
    \begin{center}
        \subfigure[FedNTD, $\beta=0$ is FedAvg]{
            \includegraphics[width=0.28\textwidth]{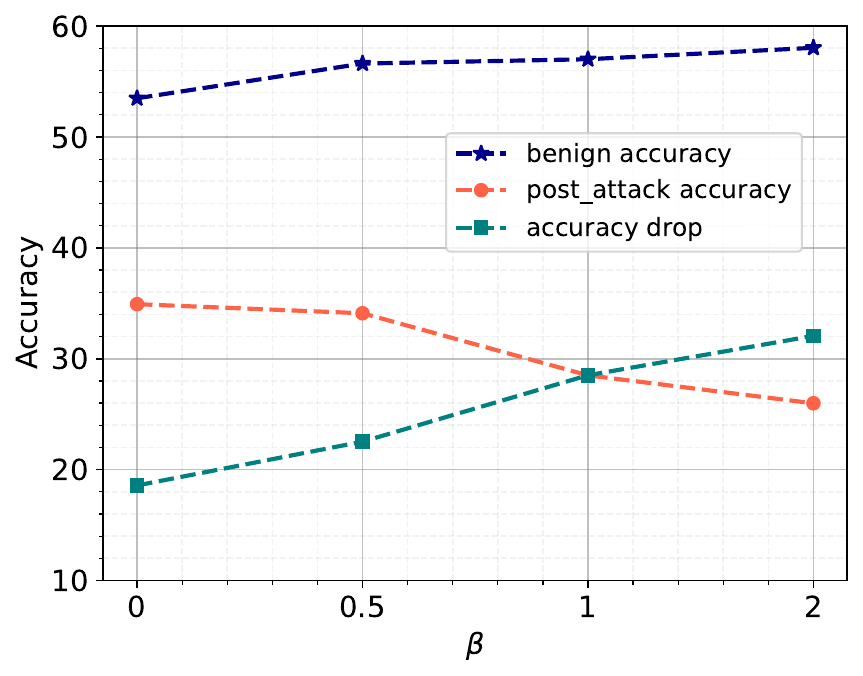}
            \label{fig:fedntd_beta_variation}
        }
        \subfigure[MOON, $\mu=0$ is FedAvg]{
            \includegraphics[width=0.28\textwidth]{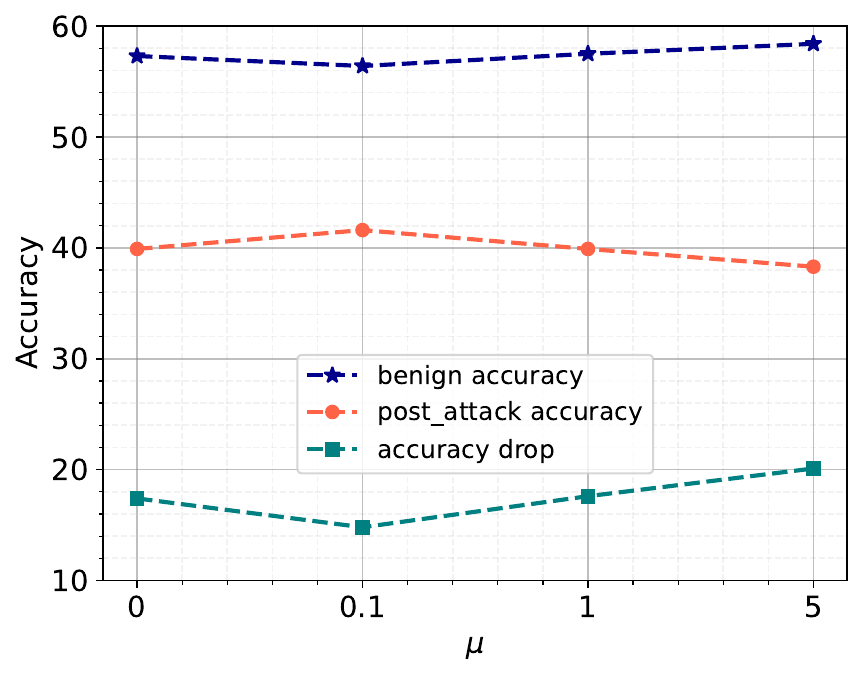}
            \label{fig:moon_mu_variation}
        }
    \end{center}
    \vspace{-1em}
    \caption{Impact of increasing KL-divergence loss for FedNTD and contrastive loss for MOON on accuracy.}
    \label{fig:fedntd_beta_moon_mu_variation}
    \vspace{-1em}
\end{wrapfigure}

\emph{Adversarial conditions.} We simulate untargeted model poisoning attacks using techniques from \cite{shejwalkar2021manipulating, fang2020local}.
To observe their effects on accuracy in both benign and adversarial settings, we vary key hyperparameters — LD-divergence loss coefficient $\beta$ for FedNTD and contrastive loss coefficient $\mu$ for MOON. The baseline for comparison is FedAvg with $\beta = 0$ and $\mu = 0$. To ensure high heterogeneity in both settings, the Dirichlet distribution~\cite{{minka2000estimating}} parameter $\alpha$ is fixed at $0.1$. 

\emph{Findings.} In Figures~\ref{fig:fedntd_beta_variation} and ~\ref{fig:moon_mu_variation}, we present three key results: benign accuracy (blue), post-attack accuracy (orange), and the accuracy drop (green). We make the following observations from increasing $\beta$ and $\mu$ are as follows: (1) the global model accuracy improves in benign settings; (2) post-attack accuracy decreases; and (3) accuracy drop increases. Our analysis shows a significant trade-off: \emph{the very mechanisms that improve performance in benign conditions (increasing $\beta$ and $\mu$) also make the models more vulnerable to adversarial attacks.}

\noindent\textbf{What causes attack amplification?}
The fundamental nature of KD-based FL methods aims to align local models with the global model. In benign scenarios, these methods significantly outperform FedAvg~\cite{li2021model, lee2022preservation}. However, in the presence of model poisoning attacks, \emph{this model alignment process inadvertently forces local models to align its representation/predictions to the poisoned global model, amplifying the attack's impact}. This is illustrated in Figure~\ref{fig:problem}, where clients \emph{unknowingly distill knowledge} from a poisioned server model.

\noindent\textbf{Formally: }
Consider a set of $n$ clients ${c_1, c_2, \ldots, c_n}$ with $m$ being malicious. Using an aggregation rule such as FedAvg, the server aggregates updates from both benign ($\nabla_{i \in [m+1, n]}$) and malicious ($\nabla^{m}_{i \in [m]}$) clients:
\begin{equation}
    \nabla_{g} = f_{\text{agr}}(\nabla^{m}_{i \in [m]} \cup \nabla_{i \in [m+1, n]})
\end{equation}
When $m=0$, the server model ${\nabla}_g^{b}$ is benign. For $m\neq0$, the server model ${\nabla}_g^{'}$ is poisoned, deviating from the ideal unpoisoned global model due to the nature of these attacks~\cite{shejwalkar2021manipulating, fang2020local, shejwalkar2022back}. Aligning local models with a poisoned global model reduces gradient diversity, making local models more similar to the poisoned global model~\cite{lee2022preservation} through KL-divergence or contrastive loss. We rewrite Equation~\ref{eqn:KD} to formalize the loss function for an FL client, using KD, where the client is the student with output $\hat{y}_c$, and the server is the teacher with output $y_s$:
\begin{equation}\label{eqn:KD_FL}
    \mathcal{L} = \mathcal{L}_{CE}(\hat{y}_{c}, y) + \beta \mathcal{L}_{KL}(\hat{y}_{c}, y_{s})
\end{equation}
Note that for the sake of derivation here, we are using $\hat{y}_c$, which represents the generic client model output. In the case of FedNTD, it can be replaced by $\Tilde{y}_c$ that represents the not-true logits of the client model, and in the case of MOON, it can be replaced by $z_{c}$ that represents the client model's high dimensional representation.

In benign scenarios, this loss function ($\mathcal{L} = \mathcal{F(\beta)}$) decreases monotonically with $\beta$ because KD brings local models closer to an unpoisoned global model. Conversely, in adversarial scenarios, it increases with $\beta$ because KD brings local models to the poisoned global model. We can write the relation of this loss function with $\beta$ as:

\begin{equation}
    \mathcal{L}(\beta) \text{ is}
    \begin{cases}
    \text{monotonically decreasing}, & m = 0 \\
    \text{monotonically increasing}, & m \neq 0
    \end{cases}
\end{equation}
Then, the derivative of the loss function is:
\begin{equation}
    \frac{d\mathcal{L}}{d\beta} =
    \begin{cases}
    < 0, & m = 0 \\
    > 0, & m \neq 0
    \end{cases}
\end{equation}
Our derivation shows that while the distillation process decreases loss in the absence of malicious clients, it increases loss in their presence, thereby leading to reduced global model accuracy. This formal analysis highlights the need for a solution that mitigates the accuracy degradation under adversarial conditions while retaining the benefits of KD under benign conditions.

\noindent\textbf{Impact of Heterogeneity.}

\begin{wrapfigure}{r}{0.65\textwidth}
\vspace{-2em}
    \begin{center}
        \subfigure[FedNTD vs FedAvg]{
            \includegraphics[width=0.3\textwidth]{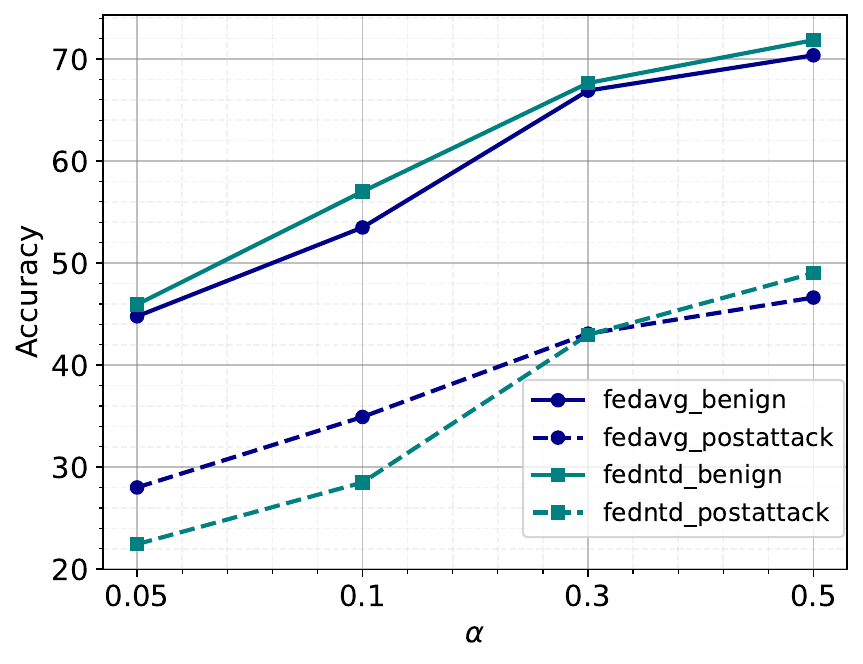}
            \label{fig:fedntd_fedavg_alpha_variation}
        }
        \subfigure[MOON vs FedAvg]{
            \includegraphics[width=0.3\textwidth]{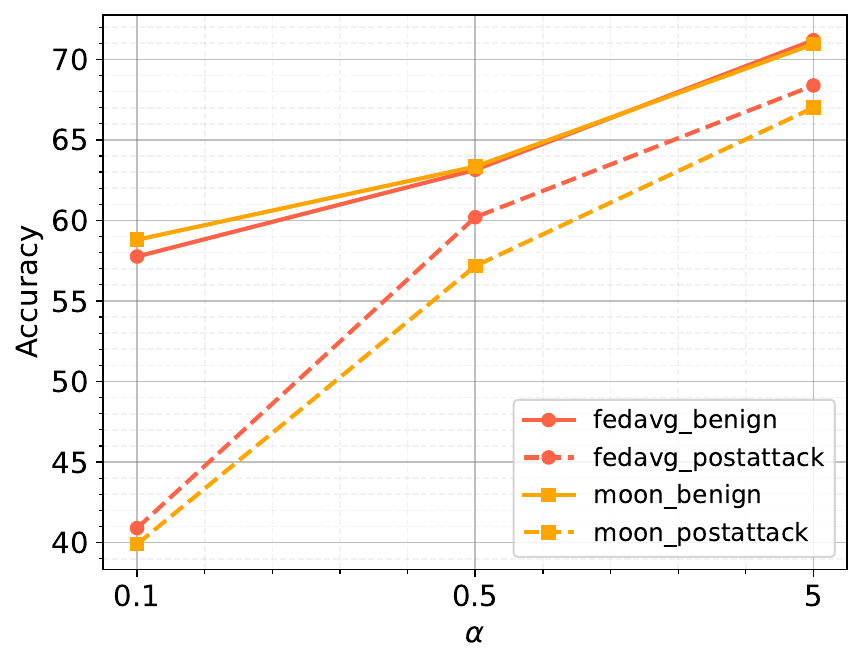}
            \label{fig:moon_fedavg_alpha_variation}
        }
    \end{center}
    \vspace{-1em}
    \caption{Impact of the heterogeneity parameter, $\alpha$ in benign and adversarial settings. We use the Dirichlet distribution where a higher $\alpha$ means lower heterogeneity.}
    \label{fig:fedntd_moon_alpha}
    \vspace{-1em}
\end{wrapfigure}
Now, we explore the effect of heterogeneity on the performance of FedNTD, MOON, and FedAvg in both benign and adversarial conditions to gain deeper insights into the role of heterogeneity in the KD performance gain vs. vulnerability tradeoff. As shown in Figure~\ref{fig:fedntd_fedavg_alpha_variation}, several interesting observations emerge. First, both FedNTD and FedAvg achieve higher accuracy at lower heterogeneity levels (indicated by higher $\alpha$). In benign settings, FedNTD consistently outperforms FedAvg. However, \emph{the trend reverses in adversarial settings:} FedAvg achieves higher accuracy than FedNTD, except at $\alpha = 0.5$. A similar pattern is observed with MOON in Figure~\ref{fig:moon_fedavg_alpha_variation}, where FedAvg outperforms MOON across all heterogeneity levels in adversarial settings. In the benign setting, as expected, MOON slightly outperforms FedAvg at high heterogeneity. This comparison highlights again how the alignment mechanisms in FedNTD and MOON with higher heterogeneity exacerbate the vulnerability of KD methods to attacks.
\section{HYDRA-FL: Hybrid Knowledge Distillation for Robust and Accurate FL}\label{sec:solution}
\begin{figure}[htbp]
    \centering
    \includegraphics[width=.8\textwidth]{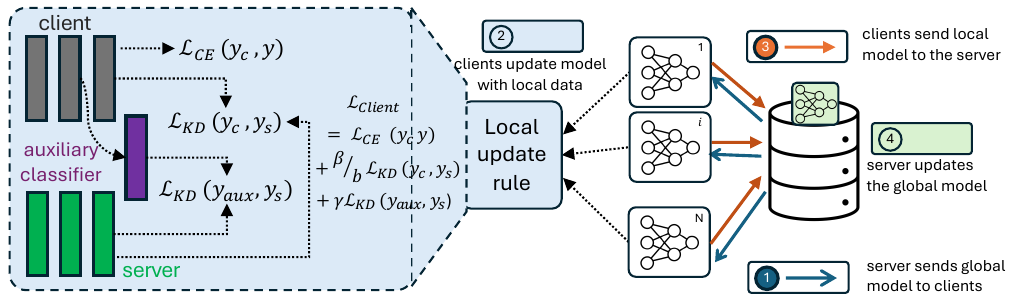}
    \vspace{-0.2cm}
    \caption{HYDRA-FL framework: we refine client model training by reducing the final layer's KD-loss and incorporating shallow KD-loss at an earlier shallow layer via an auxiliary classifier.
    }
    \label{fig:solution}
    \vspace{-0.2cm}
\end{figure}
\subsection{Generic Formulation}
In this section, we propose Hybrid Knowledge Distillation for Robust and Accurate FL (HYDRA-FL), a technique to mitigate the \emph{attack amplification} caused by KD in FL. We take a hybrid distillation approach, applying KD-loss at both the final and a shallow layer of the client model (Figure~\ref{fig:solution}). This method incorporates shallow distillation, which applies KD-loss at an intermediate layer and helps reduce the impact of poisoning by preventing over-reliance on final layer alignment. Shallow distillation previously used to handle \emph{nasty teachers} trained adversarially~\cite{kundu2021analyzing}, to reduce the impact of poisoning. In summary, shallow layers capture basic features, and shallow distillation ensures these features are robustly learned, protecting the model from adversarial influences that could corrupt deeper layers and final outputs. We first formulate the generic loss function of an FL client using KD in HYDRA-FL as:
\begin{equation}\label{eqn:generalized_CE_KD_SD}
    \mathcal{L} = \mathcal{L}_{CE}(y_{c}, y) + \frac{\beta}{b} \mathcal{L}_{KD}(y_{c}, y_{s})+ \gamma  \mathcal{L}_{KD}(y_{aux}, y_{s})
\end{equation}
This loss function has three key components:
\begin{enumerate}
    \item \textbf{Cross-entropy loss ($\mathcal{L}_{CE}(y_{c}, y)$)} is the loss between the client's prediction $y_c$ and the target $y$, drives the client model to learn from its own data, ensuring it captures \emph{in-distribution knowledge} such as features and patterns specific to its data. 
    \item \textbf{Diminished KD loss ($\frac{\beta}{b} \mathcal{L}_{KD}(y_{c}, y_{s})$)} is the loss between the client's output/representation $y_c$  and the server's output/representation $y_s$\footnote{\noindent $y_c$ and $y_s$ in generic $\mathcal{L}_{KD}$ loss can be either outputs or representations, because the method can involve either type of comparison (e.g., MOON uses representation-based loss while FedNTD has output-based loss.)}. 
    It is a strategic reduction of the KD loss to ensure that the local model benefits from the global model's knowledge while remaining robust against adversarial attacks. This approach helps balance the trade-offs between learning efficiency and model integrity. In practice, this is achieved by introducing a \emph{diminishing factor} $b$ to the KD loss at the client model's output layer to diminish the poisoning effect. The KD loss coefficient $\beta$ is divided by $b$, effectively reducing its weight in the total loss calculation, thus reducing its influence on the local model's training. This diminishing factor is essential, as shown later in \S\ref{analysis:ablation} and Figure~\ref{fig:FedNTD_ablation_beta_attack}, where reducing the $\beta$ yields better results.
    \item \textbf{Shallow distillation loss ($\gamma  \mathcal{L}_{KD}(y_{aux}, y_{s})$)} is applied at a shallow layer of the local model, enhancing robustness without heavily relying on the final layer alignment. This loss, between the auxiliary classifier's output/representation $y_{aux}$ at the client model's shallow layer and the server's output/representation $y_{s}$, is scaled by $\gamma$ to control the amount of distillation. This approach reduces the impact of poisoning on the client model. Simply reducing the KD-loss in FedNTD or MOON improves post-attack accuracy but reduces benign setting accuracy, as shown in Figure~\ref{fig:fedntd_beta_moon_mu_variation}. Our shallow distillation loss helps maintain the balance between accuracy in benign settings and lowering the impact of poisoning on the client model in adversarial settings.
\end{enumerate}

\noindent\textbf{Differences with previous works.}The key difference between our work and \cite{kundu2021analyzing} lies in our approach to shallow distillation. \cite{kundu2021analyzing} aims to distill from models that are designed to be undistillable, a.k.a \emph{nasty teachers~\cite{ma2021undistillable}}. While both use hybrid shallow distillation, \cite{kundu2021analyzing} completely removes the KD-loss from the model's output layer and uses self-distillation to compensate for performance loss due to shallow distillation. In contrast, we retain a scaled-down KD-loss at the output layer. We found that completely removing the KD-loss at the output layer may cause a more negative impact than keeping it in a reduced form.
Additionally, the untargeted poisoning is different from the poisoning in the "nasty teacher" paper~\cite{ma2021undistillable}. The "nasty teacher" performs near-perfect under normal conditions unless a malicious model distills from it. In untargeted FL poisoning, the global model is poisoned and performs poorly regardless of its use for distillation.

In HYDRA-FL, we use both final layer and shallow layer distillation to enhance robustness. \emph{Final layer distillation} aligns client outputs with server outputs for consistent predictions, whereas the \emph{shallow layer distillation} aligns intermediate representations to improve robustness against attacks. 
This dual approach reduces vulnerability to poisoning attacks, enhances learning by leveraging knowledge transfer from multiple layers, and maintains high accuracy in benign settings while being resilient under attack conditions.

\subsection{Adapting HYDRA-FL to State-of-Art Techniques}
In this section, we will adapt our generic HYDRA-FL to two state-of-the-art KD techniques for FL.

\textbf{FedNTD with shallow distillation and auxiliary classifiers.}
We modify the FedNTD base model by introducing auxiliary classifiers. The base model includes two convolutional layers, a linear layer, and a classification layer. Auxiliary classifiers, each consisting of a linear layer (hidden dimension $512$) followed by a classification layer, are added after each convolutional layer. We update the loss function to include a shallow-distillation term, representing the KL-divergence loss between the not-true logits of an auxiliary classifier and the global model. The final loss function is a weighted sum of the standard cross-entropy loss, KL-divergence loss between the not-true logits of the global model and the client model, and the KL-divergence loss between the not-true logits of the global model and the auxiliary classifier. The revised loss function in Equation~\ref{eqn:generalized_CE_KD_SD} for FedNTD is:

\begin{equation}
    \mathcal{L} = \mathcal{L}_{CE}(y_{c}, y) + \frac{\beta}{b} \mathcal{L}_{KL}(\Tilde{y}_{c}, \Tilde{y}_{s})+ \gamma  \mathcal{L}_{KL}(\Tilde{y}_{aux}, \Tilde{y}_{s})
\end{equation}
Here $y$ is the target label, $y_c$ is the client model's output, $\Tilde{y}_s$, $\Tilde{y_c}$, and $\Tilde{y}_{aux}$ are the client model's, server model's, and auxiliary classifier's not-true logits respectively.

\noindent \textbf{MOON with shallow distillation and auxiliary classifiers.}
MOON base model has two convolution layers, two linear layers, and an output classification layer. We insert auxiliary classifiers after each convolution layer. Each auxiliary classifier has two linear layers, with a hidden dimension of $256$ and an output dimension of $10$. We adapt Equation~\ref{eqn:generalized_CE_KD_SD} to MOON to compute the contrastive loss at the hidden representation layer of the auxiliary classifier as:

\begin{equation}
    \mathcal{L} = \mathcal{L}_{CE}(y_{c}, y) + \frac{\mu}{b} \mathcal{L}_{con}(z_{c}, z_{s})+ \gamma  \mathcal{L}_{con}(z_{aux}, z_{s})
\end{equation}
Here $y$ is the target label, $y_c$ is the client's output, $z_c$ is the representation from the client's final layer, $z_s$ is the representation from the server's final layer, $z_{aux}$ is the representation from the client's auxiliary classifier, and $y_s$ is the server model's output. For simplicity,  we do not write the previous round's representation in the loss function here.
\section{Experimental Results}\label{sec:experiments}

\subsection{Experimental Settings}
\noindent\textbf{Datasets and Models: }
We conduct our experiments over three popular datasets: MNIST, CIFAR10, and CIFAR100. To ensure a fair comparison with previous works, MOON and FedNTD, we utilized the same models and hyperparameters they used. Specifically, we incorporated our algorithm as a simple modification into their publicly available codes~\cite{fedntd, moon} (more details in Appendix~\ref{appdx:setup}).

\subsection{Shallow Not-True Distillation}
Our hybrid shallow not-true distillation technique significantly improves post-attack accuracy over the baseline FedNTD. As shown in Table~\ref{tab:FedNTD}, we achieve higher post-attack accuracy across all heterogeneity levels. By retaining a diminished NTD loss at the output layer, we maintain similar accuracy to FedNTD in no-attack scenarios and, in some cases, even achieve slightly higher accuracy. We also compare no-attack and post-attack accuracies for FedAvg, the foundational algorithm for many FL aggregation methods.

\begin{table}
\centering
\caption{Test accuracy for three techniques on three datasets. In the no-attack setting, (\textcolor{green}{$\uparrow$} \textcolor{red}{$\downarrow$}) shows comparison to FedAvg. In the attack setting, we use bold if our technique outperforms FedNTD.}

\label{tab:FedNTD}
\resizebox{\textwidth}{!}{%
\begin{tabular}{c||cc||cccccc||cc}
\hline
\multirow{2}{*}{\textbf{Dataset}} & \multicolumn{2}{c||}{\multirow{2}{*}{\textbf{MNIST}}} & \multicolumn{6}{c||}{\textbf{CIFAR10}} & \multicolumn{2}{c}{\multirow{2}{*}{\textbf{CIFAR100}}} \\ \cline{4-9}
 & \multicolumn{2}{c||}{} & \multicolumn{2}{c||}{$\alpha = 0.05$} & \multicolumn{2}{c||}{$\alpha = 0.1$} &  \multicolumn{2}{c||}{$\alpha = 0.5$} & \multicolumn{2}{c}{} \\ \hline
 \hline
Techniques & \multicolumn{1}{c|}{\textit{no attack}} & \textit{attack} & \multicolumn{1}{c|}{\textit{no attack}} & \multicolumn{1}{c||}{\textit{attack}} & \multicolumn{1}{c|}{\textit{no attack}} & \multicolumn{1}{c||}{\textit{attack}} &  \multicolumn{1}{c|}{\textit{no attack}} & \textit{attack} & \multicolumn{1}{c|}{\textit{no attack}} & \textit{attack} \\ \hline
Fedavg & \multicolumn{1}{c|}{92.12} & 74.48 & \multicolumn{1}{c|}{44.69} & \multicolumn{1}{c||}{31.27} & \multicolumn{1}{c|}{54.67} & \multicolumn{1}{c||}{35.67} & \multicolumn{1}{c|}{70.57} & 48.27 & \multicolumn{1}{c|}{26.17} & 12.92 \\ \hline
FedNTD & \multicolumn{1}{c|}{93.03\textcolor{green}{$\uparrow$}} & 58.09 & \multicolumn{1}{c|}{46.94\textcolor{green}{$\uparrow$}} & \multicolumn{1}{c||}{21.72} & \multicolumn{1}{c|}{56.95\textcolor{green}{$\uparrow$}} & \multicolumn{1}{c||}{32.61} & \multicolumn{1}{c|}{71.79\textcolor{green}{$\uparrow$}} & 52.51 & \multicolumn{1}{c|}{29.1\textcolor{green}{$\uparrow$}} & 13.92 \\ \hline
\rowcolor{beaublue}
HYDRA-FL(Ours) & \multicolumn{1}{c|}{92.69\textcolor{green}{$\uparrow$}} & \textbf{76.67} & \multicolumn{1}{c|}{46.92\textcolor{green}{$\uparrow$}} & \multicolumn{1}{c||}{\textbf{25.15}} & \multicolumn{1}{c|}{57.12\textcolor{green}{$\uparrow$}} & \multicolumn{1}{c||}{\textbf{34.25}} & \multicolumn{1}{c|}{71.22\textcolor{green}{$\uparrow$}} & \textbf{52.57} & \multicolumn{1}{c|}{28.9\textcolor{green}{$\uparrow$}} & \textbf{14.33} \\ \hline
\end{tabular}%
}
\end{table}

\begin{table}
\centering
\caption{Test accuracy for three techniques on three datasets. In the no-attack setting, (\textcolor{green}{$\uparrow$} \textcolor{red}{$\downarrow$}) shows comparison to FedAvg. In the attack setting, we use bold if our technique outperforms MOON.}
\label{tab:MOON}
\resizebox{\textwidth}{!}{%
\begin{tabular}{c||cc||cccccc||cc}
\hline
\multirow{2}{*}{\bfseries{Dataset}} & \multicolumn{2}{c||}{\multirow{2}{*}{\bfseries{MNIST}}} & \multicolumn{6}{c||}{\bfseries{CIFAR10}} & \multicolumn{2}{c}{\multirow{2}{*}{\textbf{CIFAR100}}} \\ 
\cline{4-9}
 & \multicolumn{2}{c||}{} & \multicolumn{2}{c||}{$\alpha = 0.1$} & \multicolumn{2}{c||}{$\alpha = 0.5$} & \multicolumn{2}{c||}{$\alpha = 5$} & \multicolumn{2}{c}{} \\ \hline
 \hline
\textbf{Methods} & \multicolumn{1}{c|}{\textit{no attack}} & \textit{attack} & \multicolumn{1}{c|}{\textit{no attack}} & \multicolumn{1}{c||}{\textit{attack}} & \multicolumn{1}{c|}{\textit{no attack}} & \multicolumn{1}{c||}{\textit{attack}} & \multicolumn{1}{c|}{\textit{no attack}} & \textit{attack} & \multicolumn{1}{c|}{\textit{no attack}} & \textit{attack} \\ \hline
Fedavg & \multicolumn{1}{c|}{88.02} & 77.55 & \multicolumn{1}{c|}{57.76} & \multicolumn{1}{c||}{40.9} & \multicolumn{1}{c|}{63.14} & \multicolumn{1}{c||}{60.2} & \multicolumn{1}{c|}{71.19} & 68.38 & \multicolumn{1}{c|}{28.36} & 24.21 \\ \hline
MOON & \multicolumn{1}{c|}{ 91.13\textcolor{green}{$\uparrow$}} & 72.32 & \multicolumn{1}{c|}{ 58.8\textcolor{green}{$\uparrow$}} & \multicolumn{1}{c||}{39.9} & \multicolumn{1}{c|}{ 63.34\textcolor{green}{$\uparrow$}} & \multicolumn{1}{c||}{57.17} & \multicolumn{1}{c|}{70.95\textcolor{red}{$\downarrow$}} & 67 & \multicolumn{1}{c|}{ 29.34\textcolor{green}{$\uparrow$}} & 23.81 \\ \hline
\rowcolor{beaublue}
HYDRA-FL(Ours) & \multicolumn{1}{c|}{92.04\textcolor{green}{$\uparrow$}} & \textbf{76.65} & \multicolumn{1}{c|}{60.1\textcolor{green}{$\uparrow$}} & \multicolumn{1}{c||}{\textbf{43.6}} & \multicolumn{1}{c|}{63.32\textcolor{green}{$\uparrow$}} & \multicolumn{1}{c||}{\textbf{59.93}} & \multicolumn{1}{c|}{70.55\textcolor{red}{$\downarrow$}} & \textbf{68.4} & \multicolumn{1}{c|}{29.48\textcolor{green}{$\uparrow$}} & \textbf{25.18} \\ \hline
\end{tabular}%
}
\end{table}

\subsection{Shallow MOON}
\begin{wrapfigure}{r}{0.65\textwidth}
\vspace{-5em}
    \begin{center}
        \subfigure[No Attack]{
            \includegraphics[width=0.3\textwidth]{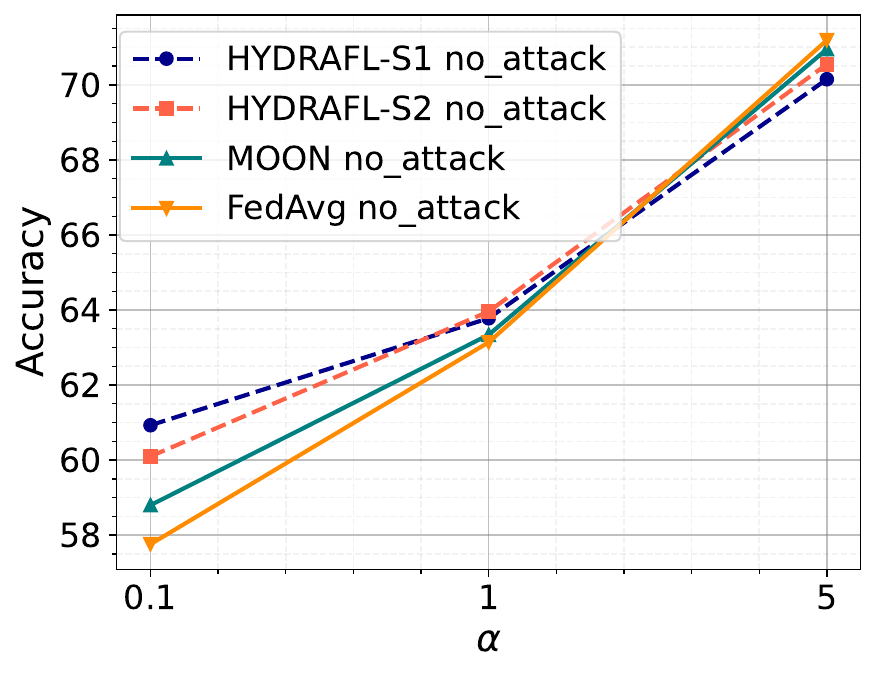}
            \label{fig:MOON_ablation_layer_no_attack}
        }
        \subfigure[Attack]{
            \includegraphics[width=0.3\textwidth]{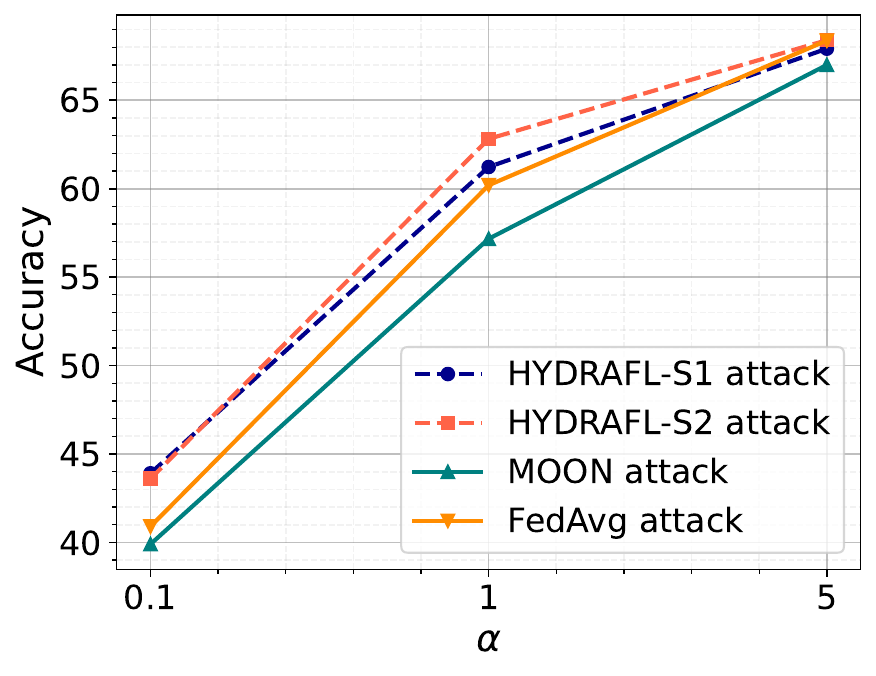}
            \label{fig:MOON_ablation_layer_attack}
        }
    \end{center}
    \vspace{-1em}
    \caption{HYDRA-FL vs. MOON and FedAvg when auxiliary classifiers are placed at different shallow layers.}
    \label{fig:MOON_ablation_layer}
    \vspace{-1em}
\end{wrapfigure}
Our shallow-distillation design effectively prevents attack amplification in MOON while maintaining nearly the same no-attack accuracy. Table~\ref{tab:MOON} shows that we achieve higher post-attack accuracy across all heterogeneity levels. Our technique also outperforms FedAvg, except in a few scenarios. Techniques like MOON are designed to enhance accuracy under high heterogeneity ($\alpha=0.1$). HYDRA-FL achieves a no-attack [attack] accuracy of $60.1 [43.6]$, surpassing both MOON ($58.8 [39.9]$) and FedAvg ($57.76 [40.9]$).
\section{Analysis}\label{sec:analysis}
In this section, we provide an in-depth analysis of HYDRA-FL. We begin with a qualitative analysis using t-distributed stochastic neighbor embedding (t-SNE~\cite{van2008visualizing}) plots to visualize the representations of the models. Then, we explore the impact of different design choices through ablation studies, focusing on the choice of the shallow layer for auxiliary classifiers and the distillation coefficients.
\subsection{Qualitative Analysis}
\begin{figure}[htbp]
    \centering
    \includegraphics[width=1.0\textwidth]{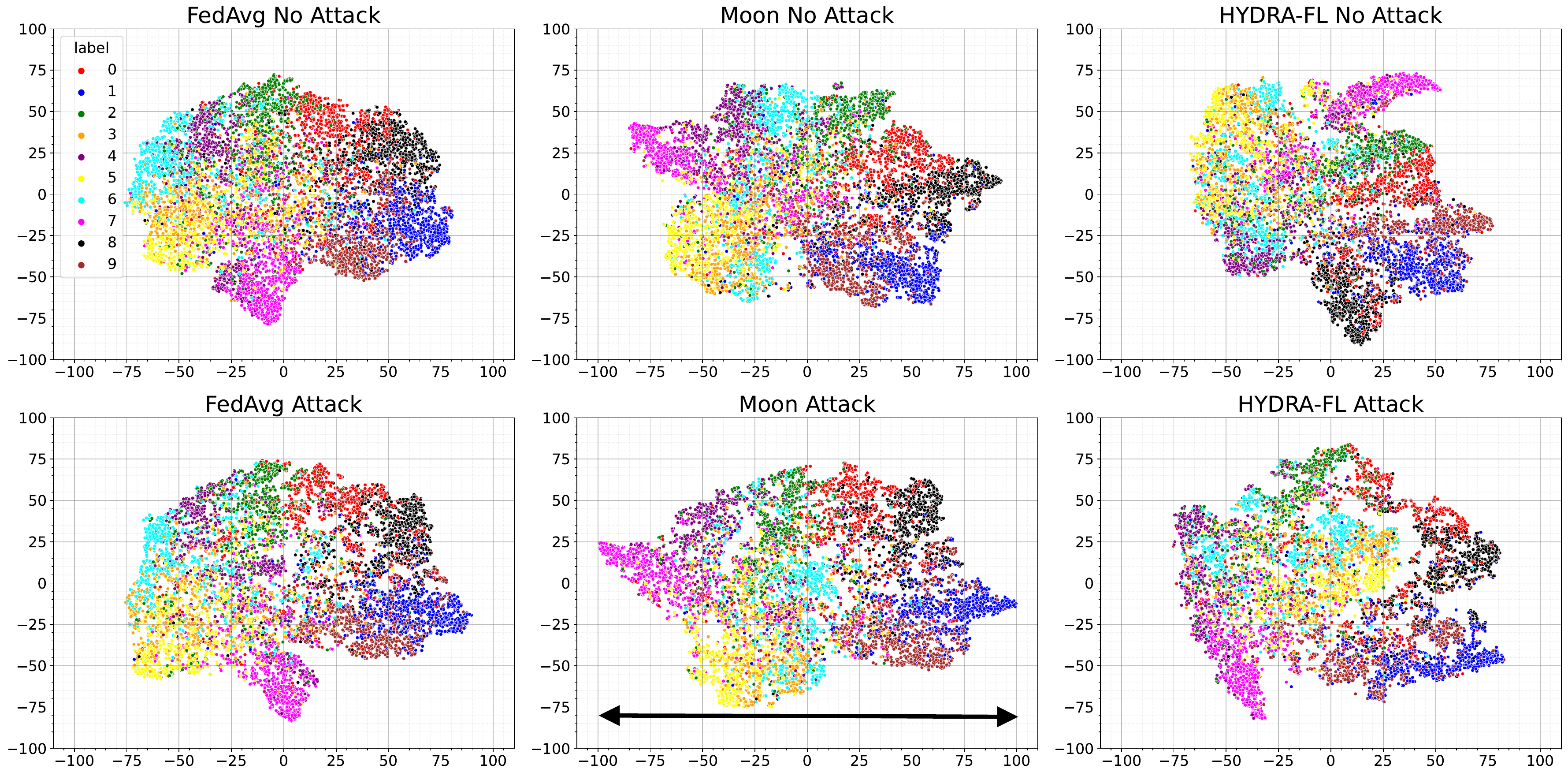}
    \vspace{-0.2cm}
    \caption{T-SNE visualizations of CIFAR10 on local model's hidden representations ($\alpha=0.5$) on FedAvg, MOON, and HYDRA-FL (ours).
    The attack vs. no-attack plot shows the deviation of the attack clusters from the no-attack clusters. Visually we can see MOON-attack has the greatest deviation, particularly along the x-axis, compared to FedAvg and HYDRA-FL.}
    \label{fig:qualitative_analysis}
    \vspace{-0.2cm}
\end{figure}

We show the t-SNE plots of the representations (Figure~\ref{fig:qualitative_analysis}) generated by the client model for FedAvg, MOON, and HYDRA-FL for both attack and no-attack scenarios. The t-SNE plots show the classes as clusters. In the MOON attack scenario, the deviation from the no-attack scenario is much higher than the deviation between HYDRA-FL with and without attack, as evident from the spread of the class clusters, especially along the x-axis.

\subsection{Ablation Study}\label{analysis:ablation}
\noindent\textbf{Impact of choice of the shallow layer.} Figure~\ref{fig:MOON_ablation_layer} illustrates the impact of the choice of the layer at which we insert our auxiliary classifier. We represent these choices by \emph{HYDRAFL-S1} and \emph{HYDRAFL-S2}, where the auxiliary classifier is inserted after the first and second convolutional layers, respectively. We compare them in both attack and no-attack settings with simple MOON and FedAvg. In Figure~\ref{fig:MOON_ablation_layer_no_attack}, both HYDRAFL-S1 and HYDRAFL-S2 outperform other techniques at low heterogeneity in the absence of an attack but slightly underperform in low heterogeneity when $\beta=5$. Figure~\ref{fig:MOON_ablation_layer_attack} shows that both HYDRAFL-S1 and HYDRAFL-S2 achieve higher post-attack accuracy at all heterogeneity levels, with HYDRAFL-S2 giving a slightly higher accuracy than HYDRAFL-S1. The benefit from the contrastive loss reduces as we go shallower, so an optimal balance is necessary.
\begin{wrapfigure}{r}{0.4\textwidth}
  \begin{center}
    \includegraphics[width=0.4\columnwidth]{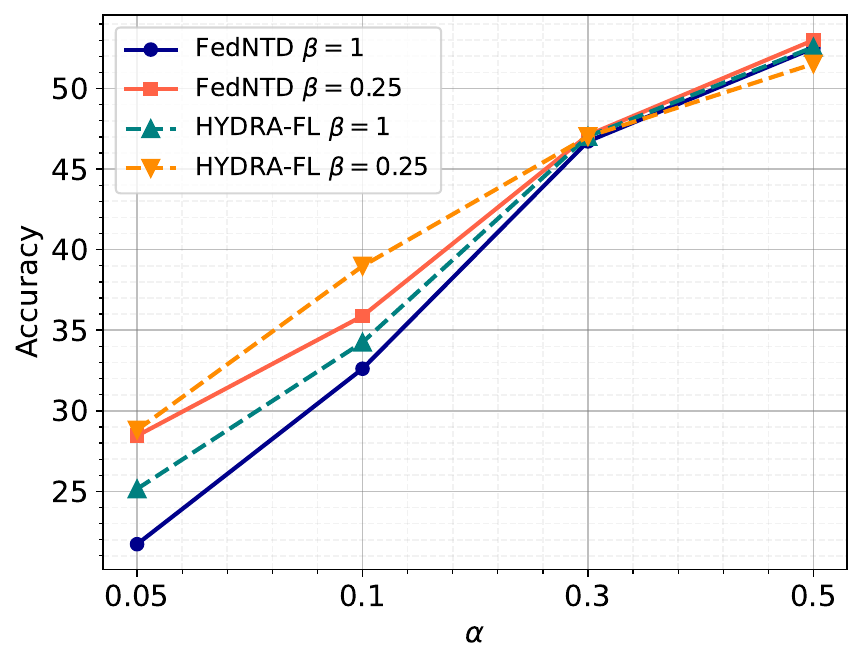}
  \end{center}
  \vspace{-1em}
    \caption{Comparison of performance of FedNTD-S with different values of $\beta$}
    \label{fig:FedNTD_ablation_beta_attack}
    \vspace{-1em}
\end{wrapfigure}

\noindent\textbf{Impact of distillation coefficients.} We examine the impact of distillation coefficients on the performance of FedNTD and HYDRA-FL. Figure~\ref{fig:FedNTD_ablation_beta_attack} shows the post-attack accuracies with two different values of the \emph{diminishing factor} $b=1,4$, resulting in output-layer NTD-loss coefficients of $\beta=1$ and $\beta=0.25$. Diminishing the coefficient $\beta$ leads to improved performance, with a significant increase in post-attack accuracy for $\beta=0.25$ at high heterogeneity ($\alpha = {0.05, 0.1}$). As demonstrated in \S\ref{sec:motivation}, $\beta$ contributes to attack amplification in FedNTD. Reducing it while performing distillation at the auxiliary classifier yields the best performance. For example, at $\alpha = 0.05$, HYDRA-FL achieves $25.15\%$ accuracy at $\beta=1$, but a much higher accuracy of $28.81\%$ at $\beta=0.25$. Similar improvements are observed at other heterogeneity levels.
\section{Conclusion}\label{sec:conclusion}

In this paper, we first identified a critical issue in KD-based FL techniques that aim to tackle data heterogeneity: in the presence of model poisoning attacks, these techniques help the attacker amplify its effect, leading to reduced global model performance. We presented empirical evidence and theoretical reasoning to back this claim. This motivated us to propose HYDRA-FL: a hybrid knowledge distillation technique for robust and accurate FL technique that aims to tackle both data heterogeneity and model poisoning, two of the biggest problems in FL. Through extensive evaluation across three datasets and comparing with baseline techniques, FedNTD and MOON, we showed that HYDRA-FL achieves superior results.


\bibliography{references}{}

\begin{thebibliography}{10}

\bibitem{technologyreviewApplePersonalizes}
{H}ow {A}pple personalizes {S}iri without hoovering up your data.
\newblock \url{https://www.technologyreview.com/2019/12/11/131629/apple-ai-personalizes-siri-federated-learning/}.

\bibitem{gboard}
Federated learning: Collaborative machine learning without centralized training data.
\newblock \url{https://ai.googleblog.com/2017/04/federated-learning-collaborative.html}, 2017.

\bibitem{bagdasaryan2018how}
Eugene Bagdasaryan, Andreas Veit, Yiqing Hua, Deborah Estrin, and Vitaly Shmatikov.
\newblock {How to backdoor federated learning}.
\newblock In {\em AISTATS}, 2020.

\bibitem{baruch2019a}
Moran Baruch, Baruch Gilad, and Yoav Goldberg.
\newblock {A Little Is Enough: Circumventing Defenses For Distributed Learning}.
\newblock In {\em NeurIPS}, 2019.

\bibitem{bhagoji2019analyzing}
Arjun~Nitin Bhagoji, Supriyo Chakraborty, Prateek Mittal, and Seraphin Calo.
\newblock {Analyzing federated learning through an adversarial lens}.
\newblock In {\em ICML}, 2019.

\bibitem{blanchard2017machine}
Peva Blanchard, Rachid Guerraoui, Julien Stainer, et~al.
\newblock {Machine learning with adversaries: Byzantine tolerant gradient descent}.
\newblock In {\em NeurIPS}, 2017.

\bibitem{cao2022fedrecover}
X.~Cao, J.~Jia, Z.~Zhang, and N.~Gong.
\newblock Fedrecover: Recovering from poisoning attacks in federated learning using historical information.
\newblock In {\em 2023 2023 IEEE Symposium on Security and Privacy (SP) (SP)}, pages 326--343, Los Alamitos, CA, USA, may 2023. IEEE Computer Society.

\bibitem{cao2021provably}
Xiaoyu Cao, Jinyuan Jia, and Neil~Zhenqiang Gong.
\newblock {Provably Secure Federated Learning against Malicious Clients}.
\newblock In {\em AAAI}, 2021.

\bibitem{chang2019cronus}
Hongyan Chang, Virat Shejwalkar, Reza Shokri, and Amir Houmansadr.
\newblock {Cronus: Robust and Heterogeneous Collaborative Learning with Black-Box Knowledge Transfer}.
\newblock {\em arXiv:1912.11279}, 2019.

\bibitem{chen2020fedbe}
Hong-You Chen and Wei-Lun Chao.
\newblock Fedbe: Making bayesian model ensemble applicable to federated learning.
\newblock {\em arXiv preprint arXiv:2009.01974}, 2020.

\bibitem{fang2020local}
Minghong Fang, Xiaoyu Cao, Jinyuan Jia, and Neil~Zhenqiang Gong.
\newblock {Local Model Poisoning Attacks to Byzantine-Robust Federated Learning}.
\newblock In {\em USENIX}, 2020.

\bibitem{hinton2015distilling}
Geoffrey Hinton, Oriol Vinyals, and Jeff Dean.
\newblock Distilling the knowledge in a neural network.
\newblock {\em arXiv preprint arXiv:1503.02531}, 2015.

\bibitem{hsu2019measuring}
Tzu-Ming~Harry Hsu, Hang Qi, and Matthew Brown.
\newblock Measuring the effects of non-identical data distribution for federated visual classification.
\newblock {\em arXiv preprint arXiv:1909.06335}, 2019.

\bibitem{kairouz2019advances}
Peter Kairouz, H~Brendan McMahan, Brendan Avent, et~al.
\newblock {Advances and open problems in federated learning}.
\newblock {\em arXiv:1912.04977}, 2019.

\bibitem{karimireddy2020scaffold}
Sai~Praneeth Karimireddy, Satyen Kale, Mehryar Mohri, Sashank Reddi, Sebastian Stich, and Ananda~Theertha Suresh.
\newblock Scaffold: Stochastic controlled averaging for federated learning.
\newblock In {\em International Conference on Machine Learning}, pages 5132--5143. PMLR, 2020.

\bibitem{khan2023pitfalls}
Momin~Ahmad Khan, Virat Shejwalkar, Amir Houmansadr, and Fatima~M Anwar.
\newblock On the pitfalls of security evaluation of robust federated learning.
\newblock In {\em 2023 IEEE Security and Privacy Workshops (SPW)}, pages 57--68. IEEE, 2023.

\bibitem{Krizhevsky2009learning}
Alex Krizhevsky.
\newblock Learning multiple layers of features from tiny images.
\newblock Technical report, University of Toronto, 2009.

\bibitem{kundu2021analyzing}
Souvik Kundu, Qirui Sun, Yao Fu, Massoud Pedram, and Peter Beerel.
\newblock Analyzing the confidentiality of undistillable teachers in knowledge distillation.
\newblock {\em Advances in Neural Information Processing Systems}, 34:9181--9192, 2021.

\bibitem{lecunmnist}
Yann LeCun and Corinna Cortes.
\newblock The mnist database of handwritten digits.
\newblock {\em http://yann. lecun. com/exdb/mnist/}, 1998.

\bibitem{lee2022preservation}
Gihun Lee, Minchan Jeong, Yongjin Shin, Sangmin Bae, and Se-Young Yun.
\newblock Preservation of the global knowledge by not-true distillation in federated learning.
\newblock {\em Advances in Neural Information Processing Systems}, 35:38461--38474, 2022.

\bibitem{fedntd}
Lee-Gihun.
\newblock Fedntd.
\newblock \url{https://github.com/Lee-Gihun/FedNTD/tree/master}.

\bibitem{li2019fedmd}
Daliang Li and Junpu Wang.
\newblock Fedmd: Heterogenous federated learning via model distillation.
\newblock {\em arXiv preprint arXiv:1910.03581}, 2019.

\bibitem{li2019rsa}
Liping Li, Wei Xu, Tianyi Chen, Georgios~B Giannakis, and Qing Ling.
\newblock {{RSA: Byzantine-robust stochastic aggregation methods for distributed learning from heterogeneous datasets}}.
\newblock In {\em AAAI}, 2019.

\bibitem{li2022federated}
Qinbin Li, Yiqun Diao, Quan Chen, and Bingsheng He.
\newblock Federated learning on non-iid data silos: An experimental study.
\newblock In {\em 2022 IEEE 38th international conference on data engineering (ICDE)}, pages 965--978. IEEE, 2022.

\bibitem{li2021model}
Qinbin Li, Bingsheng He, and Dawn Song.
\newblock Model-contrastive federated learning.
\newblock In {\em Proceedings of the IEEE/CVF conference on computer vision and pattern recognition}, pages 10713--10722, 2021.

\bibitem{li2021ditto}
Tian Li, Shengyuan Hu, Ahmad Beirami, and Virginia Smith.
\newblock {Ditto: Fair and robust federated learning through personalization}.
\newblock In {\em ICML}, 2021.

\bibitem{li2020federated}
Tian Li, Anit~Kumar Sahu, Manzil Zaheer, Maziar Sanjabi, Ameet Talwalkar, and Virginia Smith.
\newblock Federated optimization in heterogeneous networks.
\newblock {\em Proceedings of Machine learning and systems}, 2:429--450, 2020.

\bibitem{li2019convergence}
Xiang Li, Kaixuan Huang, Wenhao Yang, Shusen Wang, and Zhihua Zhang.
\newblock On the convergence of fedavg on non-iid data.
\newblock {\em arXiv preprint arXiv:1907.02189}, 2019.

\bibitem{lin2020ensemble}
Tao Lin, Lingjing Kong, Sebastian~U Stich, and Martin Jaggi.
\newblock {Ensemble distillation for robust model fusion in federated learning}.
\newblock In {\em NeurIPS}, 2020.

\bibitem{ma2021undistillable}
Haoyu Ma, Tianlong Chen, Ting-Kuei Hu, Chenyu You, Xiaohui Xie, and Zhangyang Wang.
\newblock Undistillable: Making a nasty teacher that cannot teach students.
\newblock {\em arXiv preprint arXiv:2105.07381}, 2021.

\bibitem{mahloujifar2019universal}
Saeed Mahloujifar, Mohammad Mahmoody, and Ameer Mohammed.
\newblock {Universal multi-party poisoning attacks}.
\newblock In {\em ICML}, 2019.

\bibitem{mcmahan2017communication}
H~Brendan McMahan, Eider Moore, Daniel Ramage, Seth Hampson, and Blaise Aguera~y Arcas.
\newblock {Communication-efficient learning of deep networks from decentralized data}.
\newblock In {\em AISTATS}, 2017.

\bibitem{mhamdi2018the}
El~Mahdi~El Mhamdi, Rachid Guerraoui, and Sébastien Rouault.
\newblock {The Hidden Vulnerability of Distributed Learning in Byzantium}.
\newblock In {\em ICML}, 2018.

\bibitem{minka2000estimating}
Thomas Minka.
\newblock {Estimating a Dirichlet distribution}, 2000.

\bibitem{munoz2017towards}
Luis Mu{\~n}oz-Gonz{\'a}lez, Battista Biggio, Ambra Demontis, Andrea Paudice, Vasin Wongrassamee, Emil~C Lupu, and Fabio Roli.
\newblock {Towards poisoning of deep learning algorithms with back-gradient optimization}.
\newblock In {\em AISec}, 2017.

\bibitem{pytorch}
{PyTorch Documentation}.
\newblock \url{https://pytorch.org/}, 2019.

\bibitem{moon}
QinbinLi.
\newblock Moon.
\newblock \url{https://github.com/QinbinLi/MOON/tree/main}.

\bibitem{radford2021learning}
Alec Radford, Jong~Wook Kim, Chris Hallacy, Aditya Ramesh, Gabriel Goh, Sandhini Agarwal, Girish Sastry, Amanda Askell, Pamela Mishkin, Jack Clark, et~al.
\newblock Learning transferable visual models from natural language supervision.
\newblock In {\em International conference on machine learning}, pages 8748--8763. PMLR, 2021.

\bibitem{reddi2020adaptive}
Sashank~J Reddi, Zachary Charles, Manzil Zaheer, Zachary Garrett, Keith Rush, Jakub Kone{\v{c}}n{\`y}, Sanjiv Kumar, and Hugh~Brendan McMahan.
\newblock {Adaptive Federated Optimization}.
\newblock In {\em ICLR}, 2020.

\bibitem{shejwalkar2022back}
V.~Shejwalkar, A.~Houmansadr, P.~Kairouz, and D.~Ramage.
\newblock Back to the drawing board: A critical evaluation of poisoning attacks on production federated learning.
\newblock In {\em 2022 2022 IEEE Symposium on Security and Privacy (SP) (SP)}, pages 1117--1134, Los Alamitos, CA, USA, may 2022. IEEE Computer Society.

\bibitem{shejwalkar2021manipulating}
Virat Shejwalkar and Amir Houmansadr.
\newblock {Manipulating the Byzantine: Optimizing Model Poisoning Attacks and Defenses for Federated Learning}.
\newblock In {\em NDSS}, 2021.

\bibitem{van2008visualizing}
Laurens Van~der Maaten and Geoffrey Hinton.
\newblock Visualizing data using t-sne.
\newblock {\em Journal of machine learning research}, 9(11), 2008.

\bibitem{webankcredit}
{Utilization of FATE in Risk Management of Credit in Small and Micro Enterprises}.
\newblock \url{https://www.fedai.org/cases/utilization-of-fate-in-risk-management-of-credit-in-small-and-micro-enterprises/}, 2019.

\bibitem{xie2018generalized}
Cong Xie, Oluwasanmi Koyejo, and Indranil Gupta.
\newblock {Generalized byzantine-tolerant sgd}.
\newblock {\em arXiv:1802.10116}, 2018.

\bibitem{xie2019fall}
Cong Xie, Sanmi Koyejo, and Indranil Gupta.
\newblock {Fall of empires: Breaking Byzantine-tolerant SGD by inner product manipulation}.
\newblock {\em arXiv:1903.03936}, 2019.

\bibitem{xie2022robust}
Yueqi Xie, Weizhong Zhang, Renjie Pi, Fangzhao Wu, Qifeng Chen, Xing Xie, and Sunghun Kim.
\newblock Robust federated learning against both data heterogeneity and poisoning attack via aggregation optimization.
\newblock {\em arXiv preprint}, 2022.

\bibitem{yin2018byzantine}
Dong Yin, Yudong Chen, Kannan Ramchandran, and Peter Bartlett.
\newblock {Byzantine-robust distributed learning: Towards optimal statistical rates}.
\newblock In {\em ICML}, 2018.

\bibitem{zhang2022fine}
Lin Zhang, Li~Shen, Liang Ding, Dacheng Tao, and Ling-Yu Duan.
\newblock Fine-tuning global model via data-free knowledge distillation for non-iid federated learning.
\newblock In {\em Proceedings of the IEEE/CVF conference on computer vision and pattern recognition}, pages 10174--10183, 2022.

\bibitem{zhang2019your}
Linfeng Zhang, Jiebo Song, Anni Gao, Jingwei Chen, Chenglong Bao, and Kaisheng Ma.
\newblock Be your own teacher: Improve the performance of convolutional neural networks via self distillation.
\newblock In {\em Proceedings of the IEEE/CVF international conference on computer vision}, pages 3713--3722, 2019.

\bibitem{zhang2022fldetector}
Zaixi Zhang, Xiaoyu Cao, Jinyuan Jia, and Neil~Zhenqiang Gong.
\newblock Fldetector: Defending federated learning against model poisoning attacks via detecting malicious clients.
\newblock In {\em Proceedings of the 28th ACM SIGKDD Conference on Knowledge Discovery and Data Mining}, pages 2545--2555, 2022.

\bibitem{zhao2018federated}
Yue Zhao, Meng Li, Liangzhen Lai, Naveen Suda, Damon Civin, and Vikas Chandra.
\newblock Federated learning with non-iid data.
\newblock {\em arXiv preprint arXiv:1806.00582}, 2018.

\bibitem{zhou2020distilled}
Yanlin Zhou, George Pu, Xiyao Ma, Xiaolin Li, and Dapeng Wu.
\newblock Distilled one-shot federated learning.
\newblock {\em arXiv preprint arXiv:2009.07999}, 2020.

\bibitem{zhu2021data}
Zhuangdi Zhu, Junyuan Hong, and Jiayu Zhou.
\newblock Data-free knowledge distillation for heterogeneous federated learning.
\newblock In {\em International conference on machine learning}, pages 12878--12889. PMLR, 2021.

\end{thebibliography}
\bibliographystyle{plain}
\clearpage

\appendix

\noindent\begin{LARGE}\textbf{Appendix} \vspace{4mm} \end{LARGE}

We provide additional information for our paper, HYDRA-FL: Hybrid Knowledge Distillation for Robust and Accurate Federated Learning, in the following order:

\begin{itemize}
    \item Limitations and Future Work (Appendix~\ref{appdx:limitations})
    \item Terminology/Techniques (Appendix~\ref{appdx:terminology}
    \item Adversarial Settings (Appendix~\ref{appdx:adversarial_settings})
    \item Experimental Setup (Appendix~\ref{appdx:setup})
    \item Additional Results (Appendix~\ref{appdx:additional}
\end{itemize}

\section{Limitations and Future Work}\label{appdx:limitations}

Federated Learning can have very diverse setups, especially FL in an adversarial setting. We can have many setup combinations as we can choose between different aggregation rules, attacks, defenses, datasets, data modalities, data distribution types, data heterogeneity levels, number of clients, etc. Therefore, evaluating against all combinations of these settings is well beyond the scope of one paper. Hence, for this paper, we chose only a few combinations of FL settings and tried our best to show that the problem we identified using two representative FL techniques will also exist in similar techniques. Similarly, we laid out our solution as a general framework to achieve good performance under high heterogeneity and model poisoning simultaneously. To show generalizability, we tailored it to our two representative techniques, but it would be interesting to see how our solution adapts to and performs with other FL techniques in future works. Also, we have only used unimodal, i.e., image datasets for our evaluations. This was done to stay consistent with the implementations of the techniques chosen for our case study, FedNTD and MOON. However, the language modality is becoming popular now, and multimodal models such as CLIP~\cite{radford2021learning} are being widely used as they achieve superior performance by combining both image and language modalities. We hope to incorporate language and multimodal models in our future works.

\section{Terminology/Techniques}\label{appdx:terminology}

\subsection{FedNTD}
FedNTD~\cite{lee2022preservation} is a KD-based technique that tackles the problem of data heterogeneity in FL. They first demonstrate that Data Heterogeneity causes local models to forget out-distribution knowledge, i.e., the data samples not part of the client's local data. Therefore, to preserve the out-distribution knowledge, they introduce not-true distillation, which basically modifies the loss function for the client model's local objective.
FedNTD's loss function is given by:
\begin{equation}
    \mathcal{L} = \mathcal{L}_{CE}(y_{c}, y) + \frac{\beta}{b} \mathcal{L}_{KL}(\Tilde{y}_{c}, \Tilde{y}_{s})
\end{equation}
Here $y$ is the target label, $y_c$ is the client model's output, $\Tilde{y}_s$ and $\Tilde{y_c}$ are the client model's and the server model's not-true logits, respectively.

\subsection{MOON}
MOON~\cite{li2021model} also aims to solve the problem of data heterogeneity in FL. They do so by reducing the distance between the representation learned by the local model with that of the global model. MOON's loss function is given by:
\begin{equation}
    \mathcal{L} = \mathcal{L}_{CE}(y_{c}, y) + \frac{\mu}{b} \mathcal{L}_{con}(z_{c}, z_{s})
\end{equation}
Here $y$ is the target label, $y_c$ is the client's output, $z_c$ is the representation from the client's final layer, $z_s$ is the representation from the server's final layer, and $y_s$ is the server model's output. 

\subsection{Shallow Layer and Shallow Distillation}
\label{sec:appendix-shallow}
\noindent \textbf{Shallow layer.} in a neural network refers to one of the early layers close to the input, as opposed to deeper layers that are closer to the output. In the context of a deep learning model, shallow layers generally capture low-level features, such as edges in images or simple patterns in data, while deeper layers capture more complex, abstract representations. 

\noindent \textbf{Shallow distillation.} is a technique used in KD where the knowledge transfer happens at a shallow layer of the neural network rather than at the final output layer. In traditional KD, the student model tries to mimic the teacher model's output at the final layer. In shallow distillation, an additional distillation loss is applied at one of the shallow layers of the student model. This helps the student model learn intermediate representations from the teacher, providing a more comprehensive learning experience. By aligning these intermediate representations, the student model gains a more robust understanding of the data, leading to better \emph{generalization}.

\noindent \textbf{Robustness against poisoning.} Shallow layers are less affected by adversarial attacks that target the final output of the model. Applying distillation at a shallow layer reduces the impact of a poisoned global model because the knowledge transferred is more fundamental and less influenced by the adversarial manipulations that typically affect the deeper layers.

\section{Adversarial Settings}\label{appdx:adversarial_settings}
Here we present the details of the adversarial settings of our experiments. We explain our threat model, which attacks we are using and why we are using them, and the defense we are using.

\subsection{Threat Model}\label{appdx:adversarial_settings:threat_model}
\noindent\textbf{Goal:}
Our untargeted poisoning adversary controls $m$ out of $N$ clients to manipulate the global model to misclassify all the inputs it can during testing. Unless stated otherwise, we assume $20\%$ malicious clients. Most defense works assume high percentages of malicious clients to demonstrate that their defenses work even in highly adversarial settings. Hence, although unreasonable in practical FL settings~\cite{shejwalkar2022back}, we follow prior defense works and use $20\%$ malicious clients.

\noindent\textbf{Knowledge:} Following most of the defense works, we assume that the adversary knows the robust AGR that the server uses. As assumed by most works, the adversary knows the server's AGR. To test the efficacy of our technique with a strong adversary, we consider the case where the adversary has access to not only the malicious clients' data but also the benign clients' data. This enables us to determine the upper bound of the efficacy of our technique.

\noindent\textbf{Capabilities:} Our adversary is strong enough to directly manipulate model updates of the malicious clients it controls. While poisoning attacks come in various types and flavors, we restrict ourselves to only model poisoning attacks. This is because model poisoning attacks are much stronger. It has been shown in~\cite{shejwalkar2022back} that model poisoning attacks are much stronger because they directly perturb the local model parameters. In contrast, data poisoning attacks perturb the data, subsequently perturbing the local and global models upon aggregation. Poisoning attacks can also be classified based on their error specificity. If the goal is to misclassify certain classes only, then it is a \emph{targeted attack} and is often achieved by inserting a backdoor in the model that activates only for certain inputs. On the other hand, an \emph{untargeted attack} indiscriminately lowers the accuracy for all inputs.

\subsection{Attacks we use in our evaluation}\label{appdx:adversarial_settings:attacks}
We use two model poisoning attacks for our evaluations. By testing which attack worked well, we chose the Stat-Opt attack for MOON and the Dyn-Opt attack for FedNTD. Below, we briefly explain how they work:
\begin{itemize}
    \item\textbf{Stat-Opt~\cite{fang2020local}:} gives an untargeted model poisoning framework and tailors it to specific defenes such as TrMean~\cite{yin2018byzantine}, Median~\cite{yin2018byzantine}, and Krum~\cite{blanchard2017machine}. The adversary first calculates the mean of the benign updates, $\nabla^b$, and finds the \emph{static} malicious direction $w = -sign(\nabla^b)$. It directs the benign average along the calculated direction and scales it with $\gamma$ to obtain the final poisoned update, $-\gamma w$.
    
    \item\textbf{Dyn-Opt~\cite{shejwalkar2021manipulating}:} also gives an untargeted model poisoning framework and tailors it to specific defenses, similar to Stat-Opt but differs in the \emph{dynamic} and \emph{data-dependent} nature of the perturbation. The attack first computes the mean of benign updates, $\nabla^b$, and a data-dependent direction, $w$. The final poisoned update is calculated as $\nabla^` = \nabla^b + \gamma w$, where the attack finds the largest $\gamma$ that can bypass the AGR. They compare their attack with Stat-Opt and show that the dataset-tailored $w$ and optimization-based scaling factor $\gamma$ make their attack much stronger.
\end{itemize}

\subsection{Defense we use in our evaluation}\label{appdx:adversarial_settings:defense}
We use the Trimmed Mean defense in our evaluations. Trimmed Mean~\cite{yin2018byzantine,xie2018generalized} is a foundational defense used in advanced AGRs~\cite{cao2022fedrecover,zhang2022fldetector,shejwalkar2021manipulating}. The server receives model updates from each client, sorts each input dimension $j$, discards the $m$ largest and smallest values (where $m$ indicates malicious clients), and averages the rest.

\section{Experimental Setup}\label{appdx:setup}

\noindent\textbf{Models: }
For MOON, we use a base encoder with two $5\times5$ convolutional layers, each followed by a $2\times2$ max pooling layer and two fully connected layers with ReLU activation. The base encoder is followed by a projection head with an output dimension of $256$. For FedNTD, we use a model (similar to the one in~\cite{mcmahan2017communication}) having two convolutional layers followed by a linear layer and a classification layer. For FedNTD, we test with different values and settle upon a diminishing factor $b=1$ and $\gamma$=2. For MOON, we set $\beta=0$ and set $\gamma=1$. We used PyTorch~\cite{pytorch} for our implementation on an 8GB NVIDIA RTX 3060 Ti GPU. Each run of FedNTD and MOON took about 2-3 hours on our machine.

\noindent\textbf{FL Settings: }
For FedNTD, we use 100 clients with a sampling ratio of 0.1, i.e., 10 clients are selected every round. We use momentum SGD with an initial learning rate of 0.1, weight decay of $1\times e^{-5}$, batch size of 50, and momentum of 0.9. Each run consists of 200 rounds with 5 local epochs.
For MOON, we use 10 clients with a sampling ratio of 1. We use SGD with an initial learning rate of 0.01, weight decay of $1\times e^{-5}$, batch size of 64, and momentum of $0.9$. Each run consists of 30 rounds with 10 local epochs, sufficient for convergence.

\noindent\textbf{Data Partitioning: } We use the widely used Dirichlet~\cite{minka2000estimating} distribution to generate the non-IID partitioning of data between clients. Dirichlet distribution works by sampling $p_k \sim Dir_N(\alpha)$ and assigns $p_{k,j}$ proportion of samples of class $k$ to client $j$. A lower value of  $\alpha$ corresponds to a higher level of heterogeneity since it means that most of the samples of a certain class belong to one client. Conversely, at a higher value of $\alpha$, the class samples are more evenly distributed between the clients. Also, a characteristic of the Dirichlet distribution is that both local dataset size and local per-class distribution vary across clients.

\noindent\textbf{Datasets: }The three datasets we use in our experiments are:
\begin{itemize}
    \item \textbf{MNIST~\cite{lecunmnist}: }
    MNIST is a 10-class digit image classification dataset, which contains 70,000 grayscale images of size 28 $\times$ 28. We divide all data among FL clients (100 for FedNTD and 10 for MOON) using the Dirichlet~\cite{reddi2020adaptive} distribution.
    \item \textbf{CIFAR10~\cite{Krizhevsky2009learning}: }
    CIFAR10 is a 10-class classification task with 60,000 total RGB images, each of size $32\times32$. Each class has 6000 training images and 1000 testing images. We divide all the data among 100 clients using the Dirichlet distribution, a popular synthetic strategy to generate FL datasets.
    \item \textbf{CIFAR100~\cite{Krizhevsky2009learning}: }
    CIFAR100 is similar to CIFAR10, except that it is a 100-class classification task where each class has 600 images of size $32\times32$. There are 500 training images and 100 test images per class. Like other datasets, we also partition this dataset using the Dirichlet distribution.
\end{itemize}

\section{Additional Results}\label{appdx:additional}
In this section, we present some of the additional results we have obtained.

\subsection{FedNTD}\label{appdx:additional:FedNTD}
For visual symmetry, we did not include the full table in \S\ref{sec:experiments}, but we had also run our FedNTD experiments at $\alpha=0.3$. We show the full FedNTD results in Table~\ref{tab:FedNTD-full}. Here, we can see that at at $\alpha = 0.3$ too, we achieve superior results FedAvg and FedNTD in both benign and adversarial conditions.

\begin{table}[!t]
\centering
\caption{FedNTD}
\label{tab:FedNTD-full}
\resizebox{\textwidth}{!}{%
\begin{tabular}{||c||cc||cccccccc||cc||}
\hline
\multirow{2}{*}{\textbf{Dataset}} & \multicolumn{2}{c||}{\multirow{2}{*}{\textbf{MNIST}}} & \multicolumn{8}{c||}{\textbf{CIFAR10}} & \multicolumn{2}{c|}{\multirow{2}{*}{\textbf{CIFAR100}}} \\ \cline{4-11}
 & \multicolumn{2}{c||}{} & \multicolumn{2}{c||}{\textbf{0.05}} & \multicolumn{2}{c||}{\textbf{0.1}} & \multicolumn{2}{c||}{\textbf{0.3}} & \multicolumn{2}{c||}{\textbf{0.5}} & \multicolumn{2}{c|}{} \\ \hline
 \hline
Techniques & \multicolumn{1}{c|}{\textit{no attack}} & \textit{attack} & \multicolumn{1}{c|}{\textit{no attack}} & \multicolumn{1}{c||}{\textit{attack}} & \multicolumn{1}{c|}{\textit{no attack}} & \multicolumn{1}{c||}{\textit{attack}} & \multicolumn{1}{c|}{\textit{no attack}} & \multicolumn{1}{c||}{\textit{attack}} & \multicolumn{1}{c|}{\textit{no attack}} & \textit{attack} & \multicolumn{1}{c|}{\textit{no attack}} & \textit{attack} \\ \hline
Fedavg & \multicolumn{1}{c|}{92.12} & 74.48 & \multicolumn{1}{c|}{44.69} & \multicolumn{1}{c||}{31.27} & \multicolumn{1}{c|}{54.67} & \multicolumn{1}{c||}{35.67} & \multicolumn{1}{c|}{66.34} & \multicolumn{1}{c||}{42.53} & \multicolumn{1}{c|}{70.57} & 48.27 & \multicolumn{1}{c|}{26.17} & 12.92 \\ \hline
MOON & \multicolumn{1}{c|}{93.03} & 58.09 & \multicolumn{1}{c|}{46.94} & \multicolumn{1}{c||}{21.72} & \multicolumn{1}{c|}{56.95} & \multicolumn{1}{c||}{32.61} & \multicolumn{1}{c|}{68} & \multicolumn{1}{c||}{46.72} & \multicolumn{1}{c|}{71.79} & 52.51 & \multicolumn{1}{c|}{29.1} & 13.92 \\ \hline
Ours & \multicolumn{1}{c|}{92.69} & 76.67 & \multicolumn{1}{c|}{46.92} & \multicolumn{1}{c||}{25.15} & \multicolumn{1}{c|}{57.12} & \multicolumn{1}{c||}{34.25} & \multicolumn{1}{c|}{68.1} & \multicolumn{1}{c||}{47.03} & \multicolumn{1}{c|}{71.22} & 52.57 & \multicolumn{1}{c|}{28.9} & 14.33 \\ \hline
\end{tabular}%
}
\end{table}

\subsection{MOON}\label{appdx:additional:MOON}
We also ran ablation with MNIST for different shallow layers and diminishing coefficients. We show the results in Table~\ref{tab:ablation_MNIST}, where we can see that at a lower $\mu$, i.e., higher diminishing factor, we achieve the best results. A lower $\mu$ does give us better no-attack accuracy, but we lose a lot in the attack scenario.

\begin{table}[h]
\centering
\begin{tabular}{|l|l|l|l|}
\hline
\textbf{Method} & \textbf{$\mu$} & \textbf{no-attack} & \textbf{attack} \\
\hline
HYDRA-FL s1 & 1 & 94.41 & 68.68 \\
\hline
HYDRA-FL s2 & 1 & 91.78 & 68.13 \\
\hline
\hline
HYDRA-FL s1 & 0.3 & 92.03 & 72.35 \\
\hline
HYDRA-FL s2 & 0.3 & 92.92 & 73.55 \\
\hline
\hline
HYDRA-FL s1 & 0.1 & 92.04 & 76.65 \\
\hline
HYDRA-FL s2 & 0.1 & 93.93 & 72.54 \\
\hline
\end{tabular}
\caption{Comparison of HYDRA-FL for MOON with different distillation coefficients.}
\label{tab:ablation_MNIST}
\end{table}

\clearpage

\end{document}